# Unraveling Go gaming nature by Ising Hamiltonian and common fate graphs: tactics and statistics.

Didier Barradas-Bautista[1], Matias Alvarado[*2]

1 Center of Research and Advanced Studies, IPN, Department of Mathematics, Project ABACUS, Mexico City, 07360, Mexico

2 Center of Research and Advanced Studies, IPN, Department of Computer Science, Mexico City, 07360, Mexico

* matias@cs.cinvestav.mx

## Abstract

Go gaming is a struggle between adversaries, black and white simple stones, and aim to control the most Go board territory for success. Rules are elementary but Go game fighting is highly intricate. Stones placement and interaction on board is random-appearance, likewise interaction phenomena among basic elements in physics thermodynamics, chemistry, biology, or social issues. We model the Go game dynamic employing an Ising model energy function, whose interaction coefficients reflect the application of rules and tactics to build long-term strategies. At any step of the game, the energy functional of the model assesses the control' strength of a player over the board. A close fit between predictions of the model with actual games' scores is obtained. AlphaGo computer is the current top Go player, but its behavior does not wholly reveal the Go gaming nature. The Ising function allows for precisely model the stochastic evolutions of Go gaming patterns, so, to advance the understanding on Go own-dynamic -beyond the players' abilities. The analysis of the frequency and combination of tactics shows the formation of patterns in the groups of stones during a game, regarding the turn of each player, or if human or computer adversaries are confronted.

## Introduction

Go is a two player, zero-sum and complete information game [22], that official board is a 19 x 19 grid [11]. Each player places one black/white stone on an empty board cross-point position, black plays first then white and so on. Modeling the Go gaming interaction rises similar to the modeling of the complex interaction among simple elements in nature [8, 18] and social phenomena [12, 30]. So, the mathematical modeling and algorithmic setting of Go game are meaningful in the state of the art of sciences, particularly in computer matter likewise Chess was during the 20th century [22]. A unique Go gaming challenge is to measure each player strength at any game stage. We use the Ising model [8], classic in Physics phenomena modeling, to support the design of an algorithm for quantifying the complex interaction and synergy between allied Go stones or the tension generated by the adversaries in the game. In our modeling, when a phase transition happens after a critical equilibrium state, indicates the preeminence of one player on the board.

In Go gaming, [6] white stones player receives a compensation komi by playing the second turn. Same color stones joined in horizontal or vertical line form up one indivisible compound stone. A connection of ally stones is by placing one same color stone between them. Stone's liberty is a contiguous empty board cross-point in the vertical or horizontal direction. Removal of any stone on board happens if is adversaries rounded losing all its liberties. For board territory control the way is employing tactics of invasion, reduction, nets, ladders, and connections. Stone allocation within an empty board neighborhood is an invasion, and if the adversary places a stone close to an invasion, it is making a reduction. Same color stones make a net over adversarial stones by surrounding them and make a ladder by surrounding and leaving them only remaining liberty, called Atari condition. A stone is "Go alive" if cannot be captured and is "Go dead" if cannot avoid being captured. Placement of stone being directly captured is suicide that is not allowed. Go strategies are compositions of tactics. The game ends when both players pass a turn. The score is computed based on both board territory occupied and the number of simple adversarial stones captured. The usual criteria are that the winner has the largest territorial and number of captures.

The simplicity of Go game rules makes the initial algorithmic setting simply. However, the process to attain efficient strategies is of high combinatorial complexity [3, 7]. Overcoming this complexity, historically, has been a major challenge for human Go players, and now in the XXI century, it also is for Go scientist and developers [11]. Alpha Go's decisive triumph over Lee Sedol, one of the best world human Go player in 2016, 4/5 games, was a meaningful triumph of computational intelligence [28]. Absolute preeminence of AlphaGo is the 60 - 0 simultaneous triumphs over the best human Go players in February 2017 [14, 23, 29]. The black box that complex AI represents is more difficult to understand if we consider the Go phenomenology.

This paper purpose is to advance in this comprehension. A Go gaming state is a configuration that combines black-white-empty board positions. The Go gaming state space extends with cardinality $3^{19 \times 19}$ $10^{172}$. The game tree records the different paths between the successive states that correspond to the players' decisions from the start to the end so the sequence of moves in the game. Go game tree cardinality is by $10^a$, $a = 10^{172}$, that quantifies the huge diversity of paths for Go gaming. As a result, the automation of Go tactics and strategies to efficiently win a match is vastly complex. In average, the branching factor for Go ranges from 200 to 300 possible moves at each player's turn, while 35 - 40 moves for Chess which cardinality of state space and the game tree is $10^{50}$ and by $10^{123}$ respectively [3]. As for human Go players, the hard task in Go automation is to estimate the potential to strength territory dominance for a certain play, so classify the best sequence of states picks from the enormous set of options to decide next gainful move [6, 20, 33]. Computer Go [9] uses heuristic-search [14, 29], machine learning [20] and pattern recognition techniques to identify eyes, ladders and nets [33, 35]. Monte Carlo Tree Search (MCTS) was extensively used for simulation-based search algorithms [10, 14, 15], and given a Go state, from thousands or million simulations the best average is applied to ponder the next movement [14, 15]. However, this is highly computer time-consuming. The a priori knowledge-based heuristics to identify Go tactics and strategies was added to MCTS to advance computer Go [18]. AlphaGo machine [28] uses intelligent data mining over the historical Go games to identify good Go gaming patterns. To classify these patterns and to learn from them the machine uses deep neural networks bio-inspired in animal's vision system.

The Ising model is a mathematical model of ferromagnetic properties of materials; It consists of discrete variables representing magnetic dipole moments of atomic spins that may take the dichotomy values 1 or -1. The organization of the spins is in N-dimensional lattices where each spin interacts with neighboring spins or with external magnetic fields that tend to align them in the applied field direction. This model allows the study of thermodynamic phase transitions, and the two-dimensional (2D) square-lattice Ising model is perhaps the simplest statistical model to show a phase transition. It consists in the emergence of a spin ordering, from an initial random configuration of spins pointing in either 1 or -1 direction, to a final state with spins preferentially pointing in a fixed direction, giving rise to a finite magnetization. The order emergence depends on the temperature T of the system, such that at high temperatures it is disordered while ordering manifests at temperatures smaller than a critical temperature CT. As the temperature gradually decreases below CT, ordered spin clusters first develop, subsequently percolating through the whole system, and finally leading to complete ordering at T = 0. In the 2D Ising model, the energy spin interactions are described by the Hamiltonian in Eq. 1:

$$H = \sum_{ij} w_{ij} x_i x_j - \sum_i h\, x_i \quad (1)$$

$w_i$ sets for interaction between spin *i* and *j*, the magnitude of an external magnetic field, and $h_i$ the magnetic field contribution at site *i* ; for a homogeneous external field, $h_i = 1$. Since the interaction rules in the Ising model are very general and simple [8], the approach may be applied to describe the emergence of ordering in numerous systems [19, 21, 31] in Physics, Biology, Chemistry, Sociology and technology applications [4, 25, 34], that may be assumed as constituted by discrete variables arranged in lattices and subject to extended Ising-like interacting rules. In these models, the system may develop the analog of an ordering phase transition determined by a control parameter, equivalent to the temperature in Ising models. The concept of temperature in this latter case implies the existence of thermodynamic equilibrium states. However, this property is not applicable to most systems considered in different fields. Instead, it may be assumed that the system may acquire different stationary states determined by the totality of possible configurations of its variables. Recently, Ising model was cleverly applied in nuclear medicine imaging [25], neural networks [31] and kinetics of protein aggregation studies [32], and in complex pattern recognition on biological processes to classify molecular or tissues patterns, by managing huge databases in bioinformatics [19, 27, 34]. In economy, Ising model is applied to analyze non-equilibrium phase transitions in macroeconomic modeling [30], where the emergence of patterns results from the interaction of a multitude of simple components [4]. In decision making a phase transition is a spontaneous symmetry-breaking of prices, that leads to spontaneous valuation in the absence of earnings [30], similar to the emergence of spontaneous magnetization in the absence a magnetic field.

## Materials and Methods

### Ising Hamiltonian for Go gaming

In the struggling for board area control in Go, the Ising model is relevant to modeling the dynamics of complex interaction, henceforth for designing algorithms to quantify the synergy among allied stones as well as the tension against the adversary ones. The definition of the elements of the Ising energy function helps algorithms to compute the power of stones patterns at the successive Go states, accounting for each state of dominance. After a movement brakes the black-white force equilibrium, a phase-transition-like process happens, and a dominant player in the board emerges. To efficiently enhance automation in our proposal a Go gaming state is represented by a CFG [16], see figure 1 where each Go stone is a CFG principal node, labeled with the number of single stones that compose it, and each stone's liberty is a CFG secondary node. Also, a Go gaming state representation by CFG embraces each stone's linked relationship with allies, adversaries, and liberties. By using CFG the Go sequence of moves (tactics deployment) during a game, it is easy logged, as well as, the follow up in the evolution of game interaction depicted in a lattice graph. By regarding the relative board position among allies and adversaries on the base of the CFG, this technique permits to define the Ising energy function and the design of algorithms to quantify the force of interactions among black and white atomic or molecules stones.

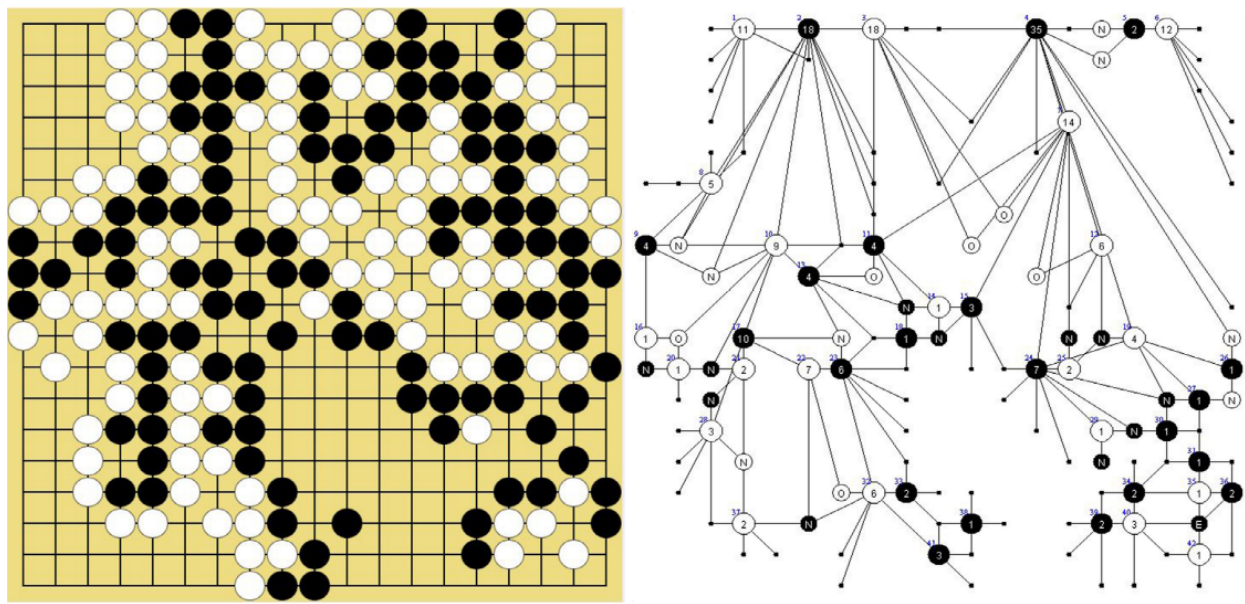

Figure 1. White stone in the left superior corner is 11 single stones, six liberties, one shared with the below white 5 single stones, and one black liberty shared with at right black 18 single stones.

## Go energy function

We use 2-dimensional Ising model for displaying the black-white stones interactions in Go gaming. Our definition of energy function directs the algorithms to compute the power of the adversary groups of stones in a Go gaming state, to observe board dominance. The energy function permits quantify the strength of interaction among allied stone, versus adversaries, and the impact of the involved liberties. Associated with Ising Hamiltonian in Eq. (1), the Go energy function via the CFG representation of states embraces the parameters mentioned in the next issues 1 and 2, then involved in the described process in issues 3 to 5:

1. The numbers of single (atomic) stones in a compound (molecular) stone.
2. The number of eyes a compound stone has.
3. The tactic pattern the stone is involved and making.
4. The synergy strength the ally stones are making among them.

5. The strength of adversary stones in the fight.

The quantitative description of stone i is employing the elements involved in Eq. 2:

$$x_i = c_i n_i + r_{eye}^{k_i} \quad (2)$$

$n_i$ sets the number of single stones, $r_{eye}$ is constant to represent the occurrence of an eye, $r_{eye} > 1$ or $r_{eye} = 0$ if no eye; $k_i$ is the number of eyes in stone $i$, and ci is the stone color, 1 for white, and -1 for black. Hence, $r_{eye}^{k_i}$ quantifies the eye's power inside $i$. If no eye $x_i$ just indicates $i$ the size and color. Observe that $k_i$ en eq. 2 guaranties these liberties to $i$, so it cannot be captured.

In Hamiltonian of Eq. 1 for Go, $w_{ij}$ should quantify the ratio of synergy or tension between single or compound stones $i, j$. So, $w_i$ should encompass the $i$-$j$ synergy regarding the presence and strength of adversary stones that try to inhibit this synergy. As well, $w_{ij}$ should encompass the presence of allied stones enforcing the mutual strengthen. Hence, up to tactics in Go gaming, the interaction among stones is weighed by the following Eq. 3:

$$w_{ij} = \sum_s r_t x_s^{ij} \quad (3)$$

$x_s^{ij}$ formula describes each stone $s$ lying between $i$ and $j$, that in turns, is making a Go tactic with allies and against adversaries.

### Tactics weighting

Parameter $r_t$ should quantify the power of each tactic $t$ : eye ($r_{eye}$), net ($r_{net}$), ladder ($r_{ladder}$), simple liberty ($r_{slb}$). The proposed values for quantify each tactic power are induced on the base of the empirical *a priori* knowledge from high rank Go human players; and, as well, on the base of the energy function definition.

Table 1 shows the assigned values for pondering each Go tactic in this work. Due to the energy function parameterization, the influence of each stone in a tactic depends on its size and its relative position on board. The usual is that more of three stones make a net and, from the middle part of the game onward, at least one stone is large, so the net power is significant. A ladder is not a frequent tactic, but its occurrence results in a strong position on the board. Besides, one compound stone may have one or two internal eyes and rarely more. Invasion is the frequent tactic for territory expansion and reduction tactic the adversary opposite move. Connection tactic results in a larger stone joining small ones; so, a connection is indirect quantified in the joined stone. Regarding these facts, the influence of each tactic is tuned in the Go Hamiltonian. For simplicity, µ = 1 in this proposal. The field impact to each stone $h_i$ is the sum of liberties the stone *i* has.

In the Go game Hamiltonian in Eq. 1 the first term accounts the interaction of collaboration among same color stones or the fight against adversaries. The second term adds the stone's force from liberties. Henceforth, given any Go gaming state, by definitions in equation 2 and 3 used in equation 1 the Go Hamiltonian allows quantifying of every stone's power. Moreover, on the base of each of them, the interaction strength among ally and/or adversarial stones: what's the contribution of each eye, ladder or net pattern; or connection tactics. Like with changes of the matter by heat or pressure transmission in natural phenomena, the evolution leading to territory control it can result in a phase transition in Go gaming. The sequentially heated stones placed as a Go move it eventually change the board state abruptly in the evolution of games, similar to matter changes. This sequence of moves yields to a Go phase-transition process that brings sudden board area dominance. In figure 2, the sequential placement of red-black-flag stones makes the override over white in this board area, similar to a local phase transition. Because Go is a zero-sum game where victory for one means defeat for the other, the Go game thermodynamics may be seen as out of the equilibrium.

| Tactic | Value |
|---|---|
| Ladder | 0.8 |
| Net | 0.6 |
| Eye | 0.4 |
| Simple liberty | 0.1 |
|  |  |

Table 1. Values for pondering each Go tactic

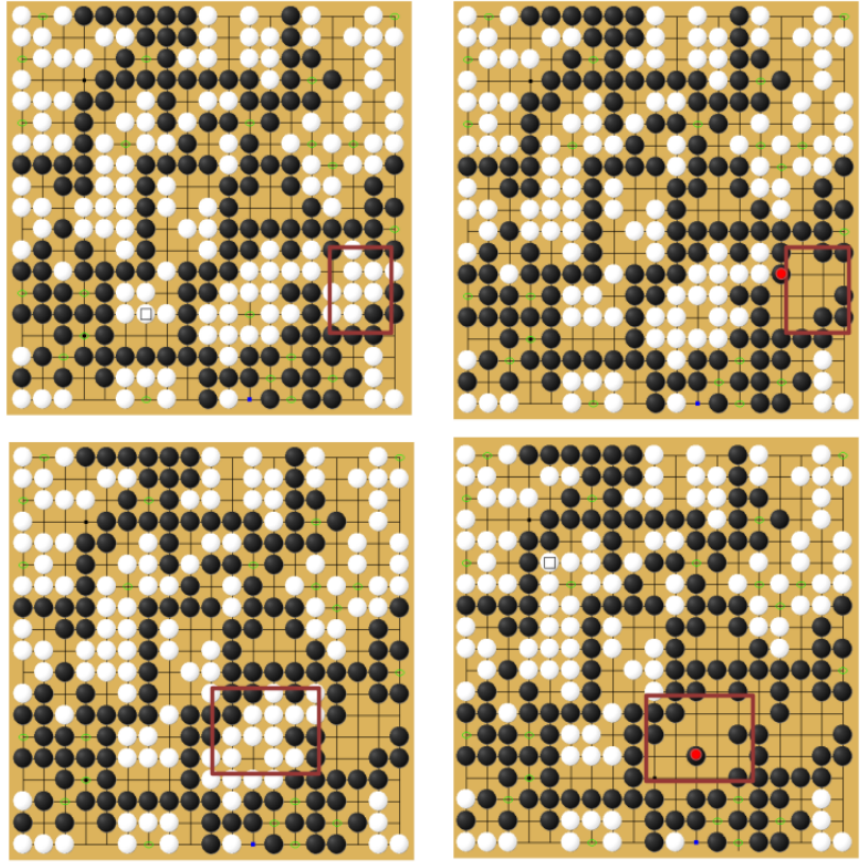

Figure 2. Go phase-transition by placement of black stone.

## Results

We use Go games' smart game files (SGF) data at http://www.go4go.net/ to test and evaluate our proposal. For experiments the SGF containing the Go game decision trees [2] is CFG translated to quantify the stones (anti-)synergy employing the Go Ising energy function. The synergy strength of black and white groups of stones pinpoint what groups are better placed on board in each state. The results of the groups of stone's strength in the states in the simulation of games are close similar to the games' score in the official tournaments [1] with top qualified Go players.

Because a Go game can be converted and stored into the SGF, several databases can sort and store the Go tournaments done with regularity during the year. We selected the badukmovies-pro-collection game data set from the Sensei website due to its clear organization by dates and games played. We analyzed more than 50 000 different professional games randomly using our in-house software. Also, in the recent year of 2017, the 60 games of Alpha Go played against human players are available on the web.

Here, we present the preference (frequency) to use Go tactics throughout different scenarios in professional games, to enlighten particularities on tactics composition that result in winning strategies.

### Black versus white in historical games

The database of Go professional games presents a framework to discover the frequency and differences on the use of tactics from the white and the black players.

Supplementary figure 1 shows the frequency of use of the different tactics in the time frame of the moves/plays of players in games we named historical, between top human Go masters mainly in the XX century. We analyzed the distribution for a preferred tactic using density plots of this games. Comparing the count of tactics and density plots of black and white player we did not observe apparent differences between their respective use of tactics. In fact, the overall distribution of the different tactics resembles a normal distribution (Supplementary figure 1). The density plots from the resulting distributions have almost the same shape with two possible populations of the tactics that are distinguished and land out of the average use (Supplementary figure 2).

### Black versus white in AlphaGo games

We counted the number of tactics used during the games by each move/play done by the players (Supplementary figures 3), black or white; regardless it was AlphaGo or a Human. The overall count revealed that net and ladders are by far the most used tactics. We observed small differences between both players, the most notorious is that the line showing the used ladder tactics has a flatter plateau in the white players. Also, the line of net tactic has a steep ascend until the play 125, and from there, the fall present three clear plateaus in its use from white, that is not so readily clear from the black player. To analyze and better visualize the differences between the tactics used we calculated the frequency of use and plotted it against its Z score. The frequency of the use of tactics also shows differences between black and white stones from AlphaGo games.

To explore the patterns in the 60 games played by AlphaGo versus humans, we used density plots to observe the distribution of the usage of all the different tactics. Again, the densities of the used tactics were above the overall count of AlphaGo and Humans mixed and classified as black or white players. The overlapping of the distributions shows small difference between the tactics population between black and white players. This behavior is expected as each player usually respond to the opponent moves accordingly to neutralize the invasive effect, especially deep into the game. In this case ladder and nets present two distinct populations of tactics that are employed much more than the mean reflected in the high value of the Z score. The respective differences in the frequency of ladder and simple liberty tactics are small. We should observe that the simple liberty tactics are associated with complex tactics, and the liberties can help to expand ladders and nets late influence (Supplementary figure 4).

## AlphaGo versus human top Go players

We analyze the 60 games played by AlphaGo against human top world Go players. At first glance, the graphs on the count and distribution of used tactics by AlphaGo or human Go players show the same shapes. However, in the count graph we observe that the games were shorter, 50 moves in average, when the AlphaGo machine plays black than when it plays white (figure 3).

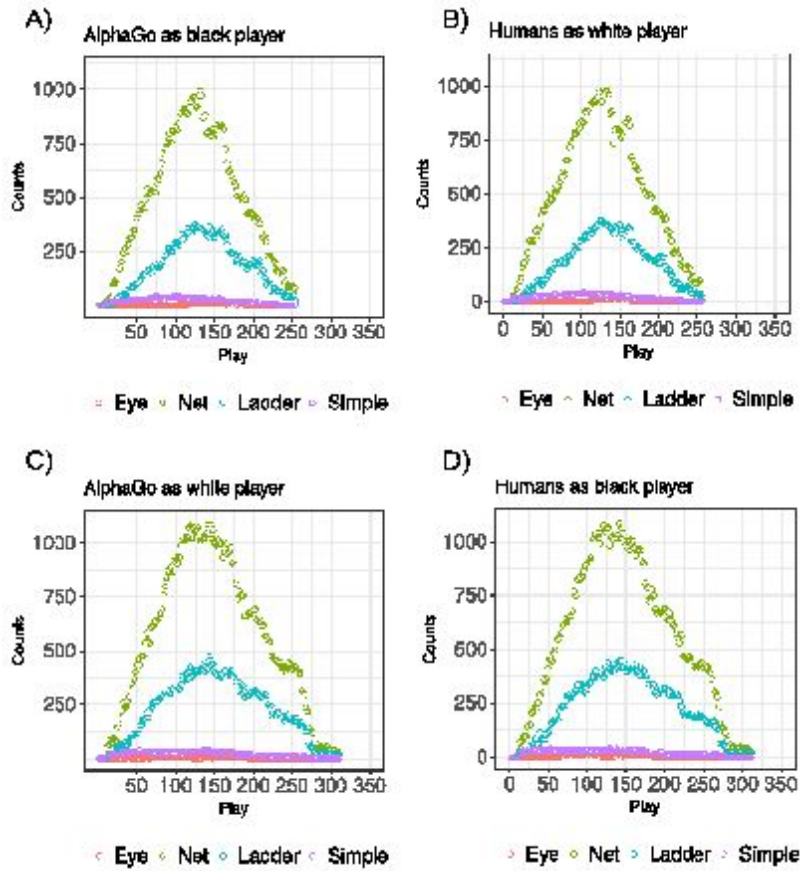

Figure 3. Scatter plots showing the number of times a strategy used as the different games progresses. A)AlphaGo as black player. B)Humans as white player. C)AlphaGo as white player. D)Humans as black player. The x axis shows the moves or plays in the games analyzed.

The shorter duration of the games is the first clear difference in the density shape of all tactics. Using the frequency distribution enables a closer inspection revealing that AlphaGo style of play has other notorious differences concerning humans' style. The density plot shows that when AlphaGo plays black, the use of eyes is very infrequent having only two population that is slightly distant from the mean use of this tactic; when AlphaGo plays white, the two populations of eye tactic are bigger than the eye tactic population from human players. The use of the ladder tactic shows a decreasing trend with only one population being more used than the average. This last fact contrasts with the use of the net tactic that increases and presents two population very distant from the average use (figure 4).

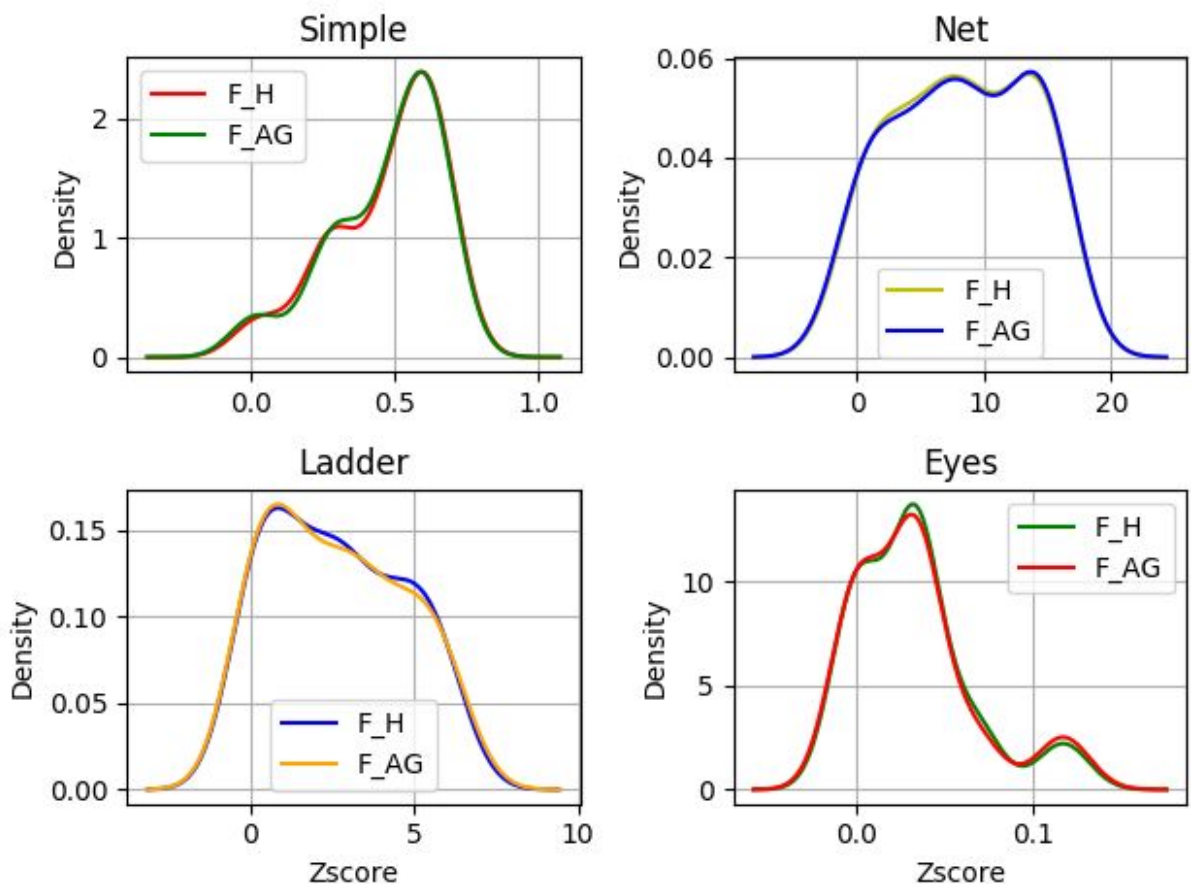

Figure 4. Density plots of the frequency of distribution against the Z score for the tactics used by AlphaGo as black player (F A G) and Humans as white player(F H ) in the 60 game tournament.

Furthermore, in close inspection, AlphaGo also present a wider Z score of the density for the ladder tactics use than human players made. We would say that the patterns of use of the eye, ladder and net tactics are related to the final victory of AlphaGo over the human players (figure 5).

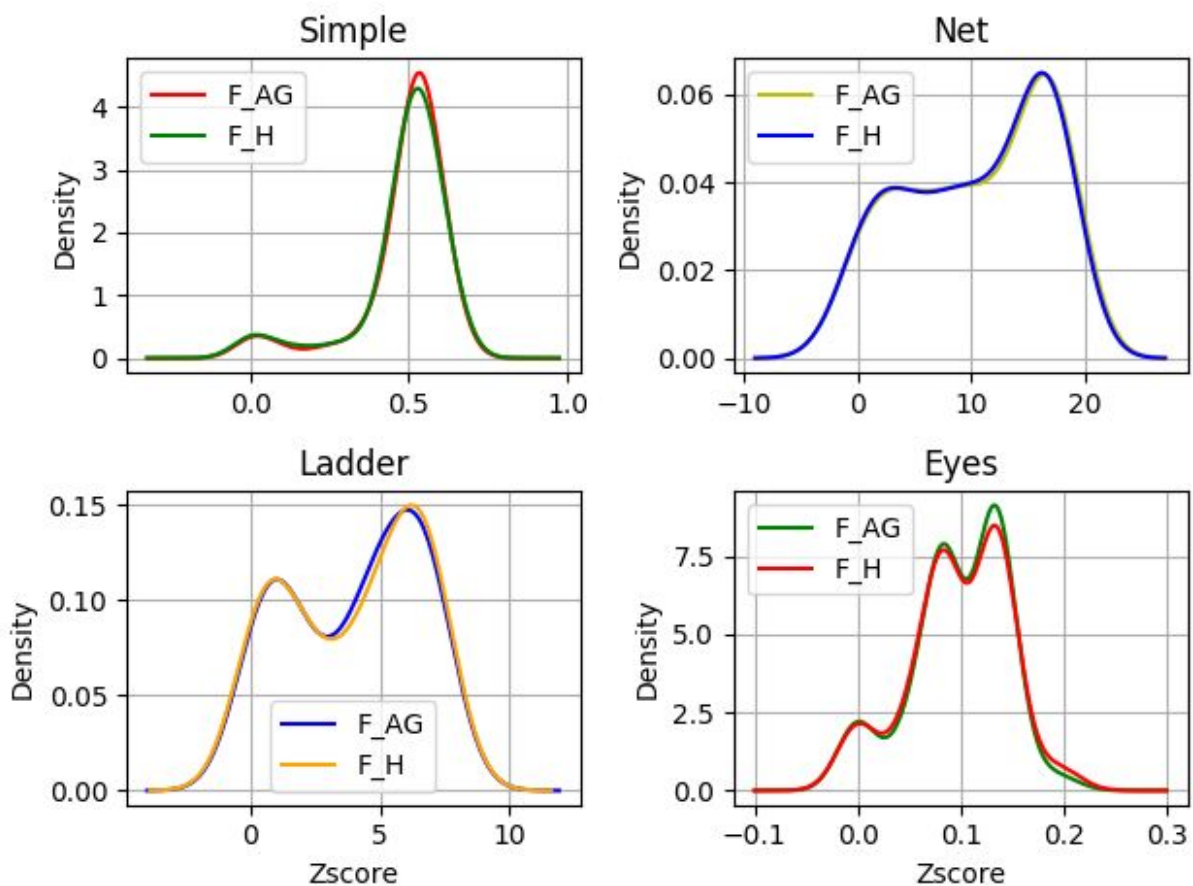

Figure 5. Density plots of the frequency of distribution against the Z score for the tactics used by AlphaGo as white player (F A G) and Humans as black player(F H ) in the 60 game tournament.

Beyond the apparent differences between AlphaGo as the black or white player, the application of the Ising model to deal with the dynamic of the Go game proved to be more informative (figure 6). The successive Ising energies of the black player show from the start phase changes, even more clearly after the 50th play/move. The most extreme phase change happens just before the 150th move/play ending with a broad energy difference around the 200th move/play (figure 6A). The Ising energy when AlphaGo plays white seems to be more compact. This could mean that the human that plays black has the advantage starting the game, so enables him to keep a balance on the board, also keeping the opponent at bay. However, after the 200th move/play abrupt phase changes in a very short span of moves leading to AlphaGo sizing the territory (figure 6B). We also observed the change in the Ising energy of only white player or black player. However, no trend could be observed (Supplementary figure 5).

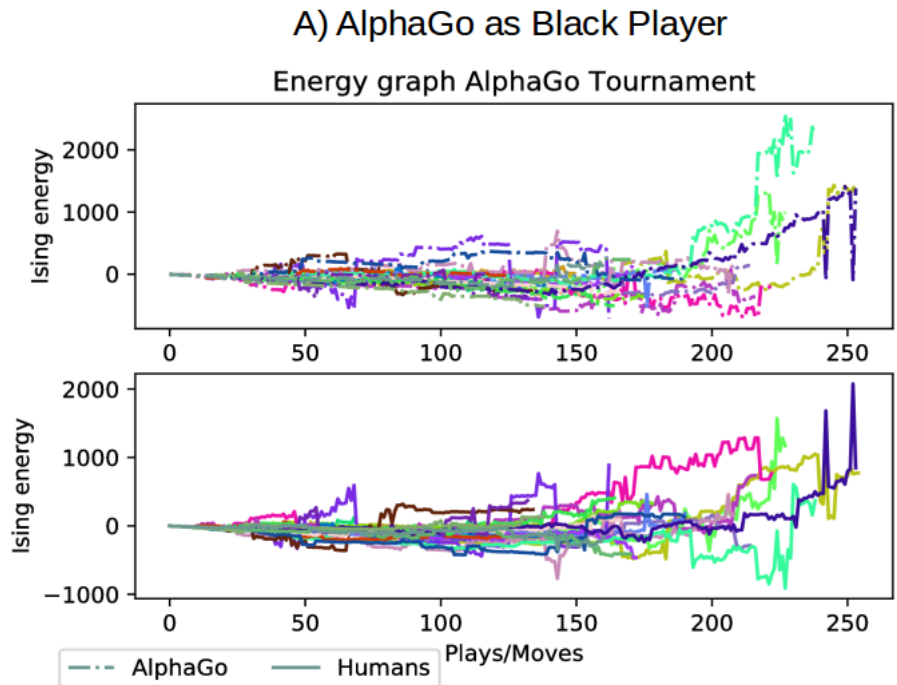

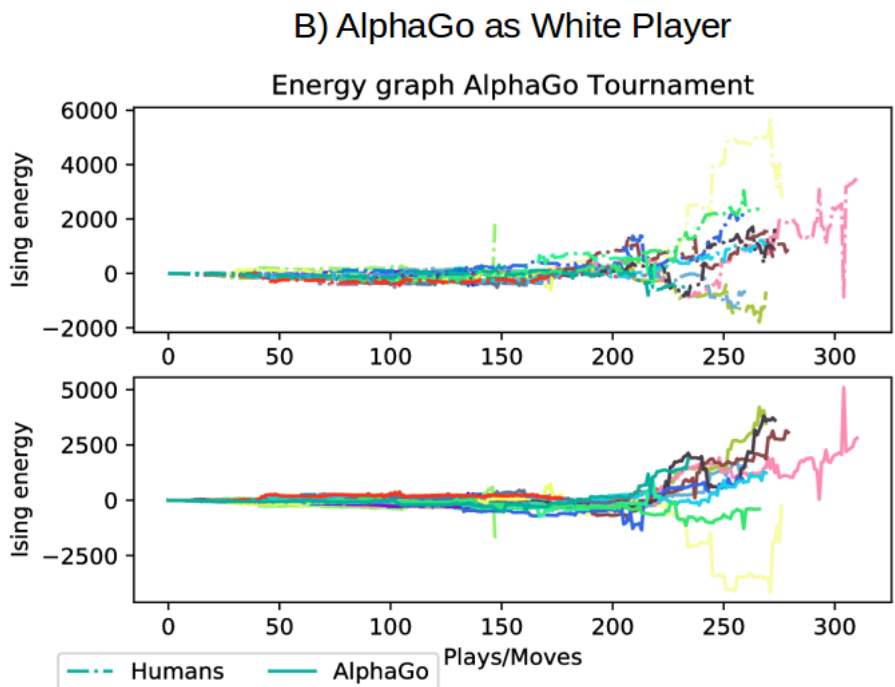

Figure 6. The Ising energies of the games in the AlphaGo tournament. A) AlphaGo plays black and start the game. B) AlphaGo Plays whites and goes after Humans start the game. The y axis shows the Ising energy of the encounter, also allowing to see the phase changes in the game where the overall energy of the math changes in favor of one of the players. The X axis shows the moves or plays for each game.

## Human versus human Go games

As a next test we simulate Go games among top human players only. In the following figures, the blue line is for the black player and red line for the white player. We made hundreds of experiments applying the Ising energy function to evaluate the strength of groups of stones on board, from initial to final states, in games reported in http://www.go4go.net/. The quantification made for some games are shown in figure 7 and in supplementary figure 6. In the figure 2 the board shows the game final state. We consider that within the same rank, Go adversaries, black and white triumphs have a normal distribution like the results of flipping a coin, so the Central Limit Theorem applies. Hence, we broadly analyze 30 games as representative. There, we got scores close to the real ones of games: in 19/30 games we fit entirely; in 8/30 games got a minor error than 5%; in the other 3/30 games, results have a range of error within 5% -10%. Please, see supplementary material: additional figures, SGF (smart game files) for simulations, and Go application for testing results in http://delta.cs.cinvestav.mx/~ matias/TeoriaJuegos/Go/PhA/principal.html

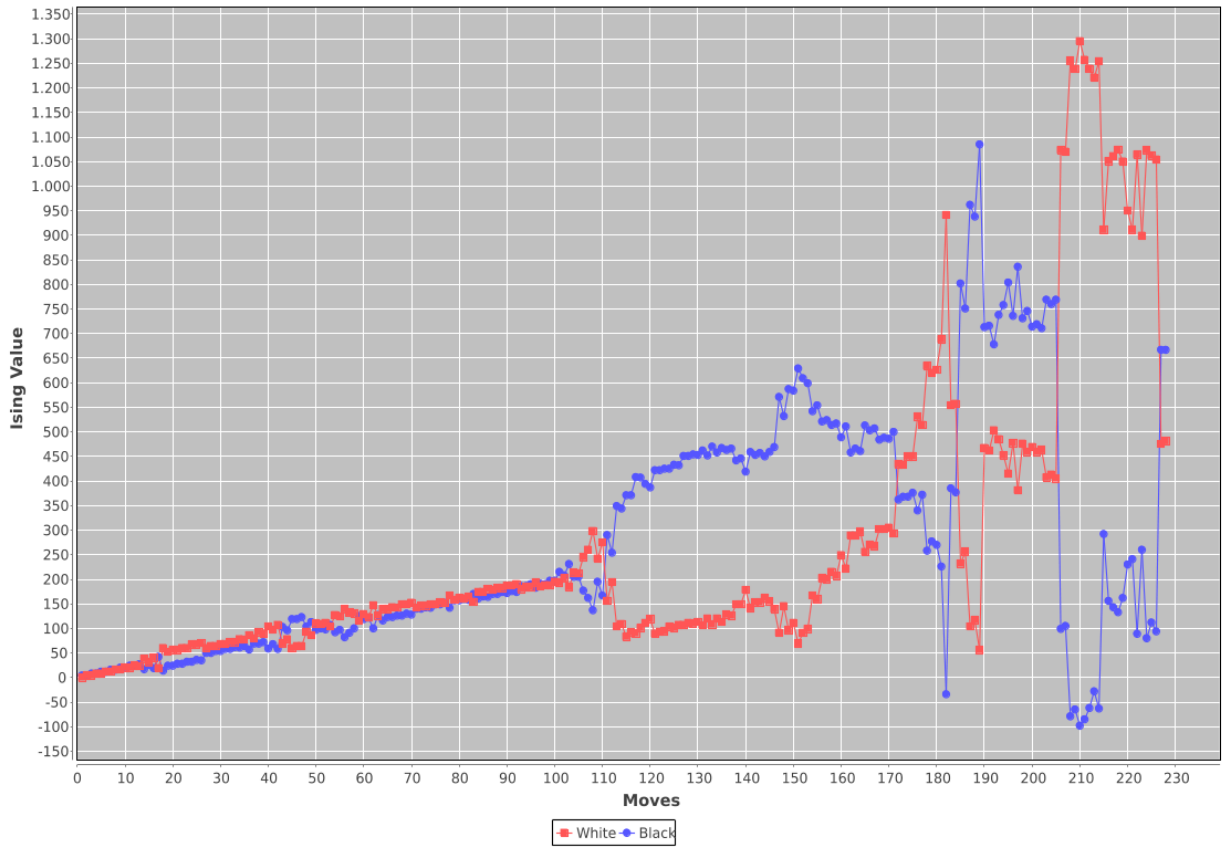

Figure 7. Murakawa plays black stones versus Cho Chikun with whites in the 39th Japanese Kisei. Until move 100 the stone's strength is tied. From move 100 to 170 the black strength is better, then phase transition and from move 171 to 230 and black strength improves. The final black strength is 667 and 481 for white. The successive reported scores are similar and the victory is for black.

### AlphaGo versus Lee Sedol games

Lee Sedol is one top Go player since he was 12 years old. AlphaGo { Lee Sedol encounter was a five-game match gamed by March 2016 in Seoul, Korea. Lee Sedol played blacks the odd games and whites the even games. AlphaGo won the first, second, third and fifth game and Lee Sedol the fourth, so 4/5 games won AlphaGo, being the first time a computer defeated a top master. Figure 8 shows the graphs of each of the five games: A, B, C and E, each score similar to the official result. In C there is a small difference between the official result and the obtained from our simulation.

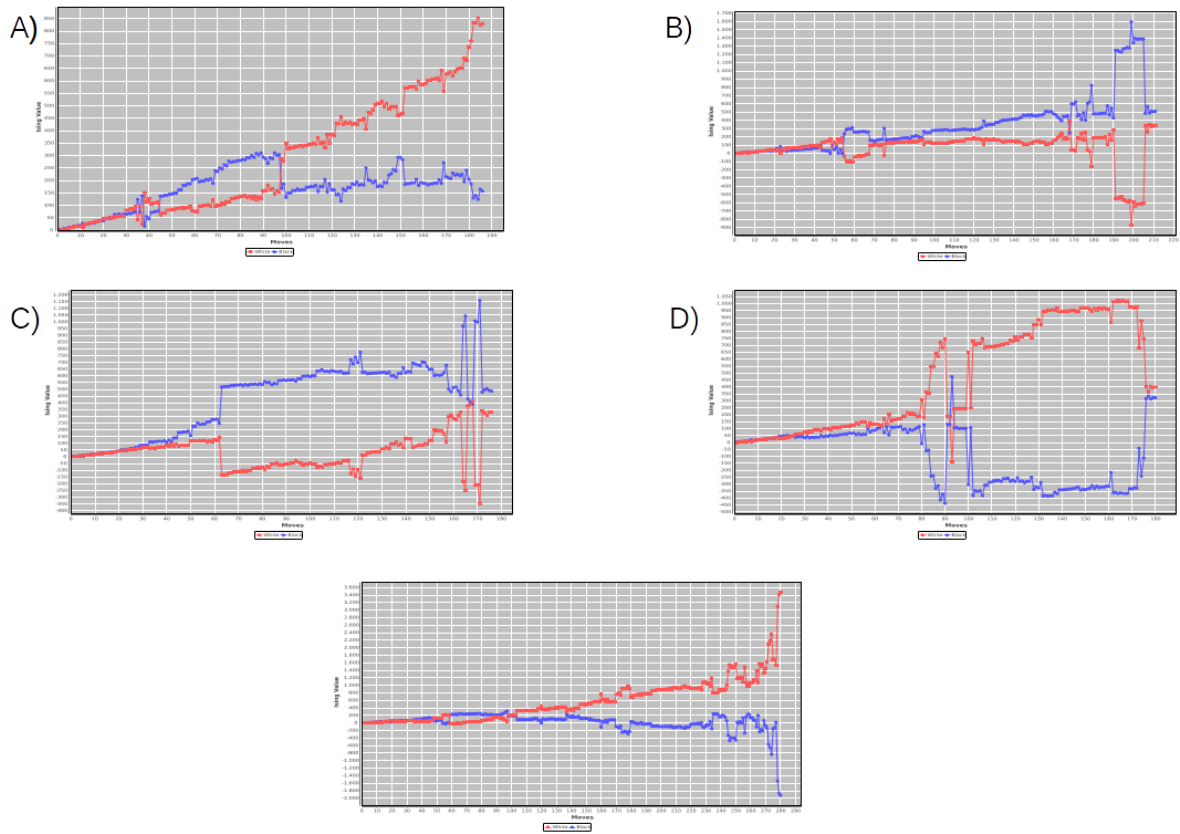

Figure 8. Ising analysis of the five games played between Lee Sedol vs. AlphaGo.

## An AlphaGo remark

The ability of AlphaGo gaming to keep a kind of self-equilibrium is quite notorious. The algorithmic design of AlphaGo makes it avoids the temptation to try an ever-absolute board control. In the AlphaGo gaming style is quite enough to keep its current advantage, and gives up to certain board area influence by predicting its strength over the whole board and game. This feature contrasts with the human Go player's behavior trying to get the absolute prevail over the adversary. Human Go players usually play to get an absolute dominance, and by this mindset, the human player loses details of weak positions, that eventually leads to losing points or the whole game. Obsession with ever increasing board control is not a weakness of AlphaGo as a remarkable quality. This virtue entails an exceptional knowledge of the game board configuration at any play made.

## Discussion

Go gaming is long-term influence moves [14] so the relevance to right play the early moves that strongly affect the outcome of late ones. Historically, in ancient cultures, Go games spent months, and most of this time during the ten first moves on average. Current competition long hours, and most of the time is spent in the first dozen of the moves as well. The AlphaGo major advance in Go automation is the bio-inspired method that emulates the visual cortex of felines and eagles characterized by an acute vision system. It results in the artificial deep neural network (DNN) [28] that embraces dozens of layers each with a not short number of neurons; the strong correlation and composition, like in visual cortex process, makes DNN a top tool for recognition in complex scenarios. AlphaGo DNN formalism uses convolution integral functions for neurons activation [28]. The DNN skills for recognition and discrimination of meaningful perceptual stimuli it presupposes the active formation of stable perceptual elements to be recognized and discriminated. Even the computer AlphaGo's ultimate purpose was to apply the best training for an artificial Go player regardless the game nature understanding. Go Ising energy function precisely modeled the stochastic evolutions of the stone patterns, step by step constructed by Go tactics during a game. The function included the atomic or molecular elements and how each other influence through time is passing and temperature changes. The stronger the interaction among same color stones the weaker the interaction among adversary ones and the dynamic in Go gaming converges to phase transitions phenomenon, furthermore interaction in natural phenomena in physics, chemistry, and biology, all of them modeled by Ising model. Go gaming temperature comes from the player's strategic talent displayed during the game. The closeness between our algorithmic simulation's results and the scores from games in top Go tournaments makes relevant our claim that comprehension on the phenomenology of Go gaming is advanced by applying the Ising model to analyze the evolution of the interaction patterns during a game. Regardless the Ising model energy function is not defined to support the Go player's abilities it makes a fair assessment of current game state. We are working on the application of the Ising-model-based algorithms to suggest a next good play. The statistical analysis might help to identify a sequence of tactics to make effective strategies in this stochastic game process. In game theory, the interaction of the spin-player and the state-action in the Ising model is for payoff and apply the Nash equilibrium [13], to analyze repeated games [36], or to modeling systems on nearest neighbor interaction with phase transitions [26]. In evolutionary dynamics of cooperation in the prisoner's dilemma game, results in a dipole-model-like, interpreting the start of a lattice of cooperation as a thermodynamical phase transition [12]. Usually, in a game, the Nash equilibrium fixes the strategy choices to lose no more than the others; so, Nash equilibrium may allow avoiding a Go phase transition that would occur if the right strategy is not applied. The computer (cost) complexity for processing a problem solution [5], refers both, the time or number of execution steps, and space or memory amount an algorithm uses too. Go gaming complexity is EXPTIME-complete [17], and more precisely PSPACE-complete [24]. The stochastic feature of Go gaming captured by the Ising-Hamiltonian-based algorithms suggests that, on this base, low-cost algorithms (a lot of less computer time and memory) could be deployed to identify the relevant strategies for success.

## Conclusions

The highly intricate Go gaming for territory control is formally traced by the Ising energy function, hence, the application of Go rules and tactics for building complex strategies that emerge in successful patterns. Beyond the usefulness to predict real games' scores, the Ising-based computer simulations advance on reveals the Go gaming nature (phenomenology): as dichotomy variables interaction process -beyond the player's abilities- that dynamically change regarding the relative positions among ally or adversary stones. From the frequency and combination of tactics the patterns formation are identified, either by regarding the each player's stones color, or the characteristic style of a human or computer player. The stochastic interaction between simple Go stones regarding elementary rules too, is close similar to (Ising) modeling the interaction phenomena among basic elements proper of natural sciences, or abstract typical entities in social sciences. Thus, advance in the comprehension of emerging complex behaves from stochastic interaction among basic entities is achieved.

## Author contributions statement

M.A conceived the use of Ising Hamiltonian for Go modeling, and conceived the experiments, D.B.B conducted the simulations and experiment(s), D.B.B and M.A analyzed the results and reviewed the manuscript.

## Additional information
### Competing financial interests
The authors and the funding agencies declare no competing interest

# Supporting Information

## *Acknowledgments*

To Carlos Villarreal Lujan (UNAM Physics Institute) for his clever advice on the use and meaning of Ising model and Hamiltonian. To Ricardo Quintero Zazueta, world Go senior master who suggested test simulations with the classic games. This work was partially supported by ABACUS, CONACyT grant EDOMEX-2011-C01-165873.